\documentclass[twoside,11pt]{article}
\usepackage{jmlr2e}

\usepackage{url}

\begin{document}

\title{How good is the Electricity benchmark for evaluating concept drift adaptation}

\author{\name Indr\.e \v{Z}liobait\.e \email zliobaite@gmail.com\\
       \addr Bournemouth University\\
       Poole, BH12 5BB, UK}


\maketitle

\begin{abstract}
In this correspondence, we will point out a problem with testing adaptive classifiers on autocorrelated data. 
In such a case random change alarms may boost the accuracy figures.  
Hence, we cannot be sure if the adaptation is working well.
\end{abstract}


\section*{Testing on the Electricity data}

\emph{Concept drift} \citep{Widmer96} has become a popular research topic over the last decade and a lot of adaptive classification algorithms have been developed.
The setting is as follows. Multidimensional input data is arriving over time, the goal is to predict the class label $y$ for each instance $X$.
In the stationary settings a classifier ${\cal L}: X \rightarrow y$ is trained once and remains fixed during the operation.
In the concept drift scenario the joint data distribution, i.e. $p(X,y)$, may change over time, and, as a result, a fixed predictor ${\cal L}$ may lose accuracy over time.
In the concept drift scenario a classifier can be updated at every time step taking into account the most recent data:
${\cal L}_t = f({\cal L}_{t-1},X_{t-1},y_{t-1})$, where $t$ is the time step, $f$ is the adaptation function.

The Electricity dataset due to \citet{Harries99} is a popular benchmark for testing adaptive classifiers\footnote{The dataset is available from e.g. \url{http://www.liaad.up.pt/~jgama/ales/ales_5.html}.}.
It has been used in over 40 concept drift experiments\footnote{Google scholar, 2013 January}.
The dataset covers a period of two years ($45\,312$ instances recorded every half an hour, 6 input variables).
A binary classification task is to predict a rise (UP) or a fall (DOWN) in the electricity price in New South Wales (Australia).
The prior probability of DOWN is $58\%$.
The data is subject to concept drift due to changing consumption habits, unexpected events and seasonality.

\begin{figure}
\begin{minipage}[b]{0.5\linewidth}
\centering
\includegraphics[width = 0.95\textwidth]{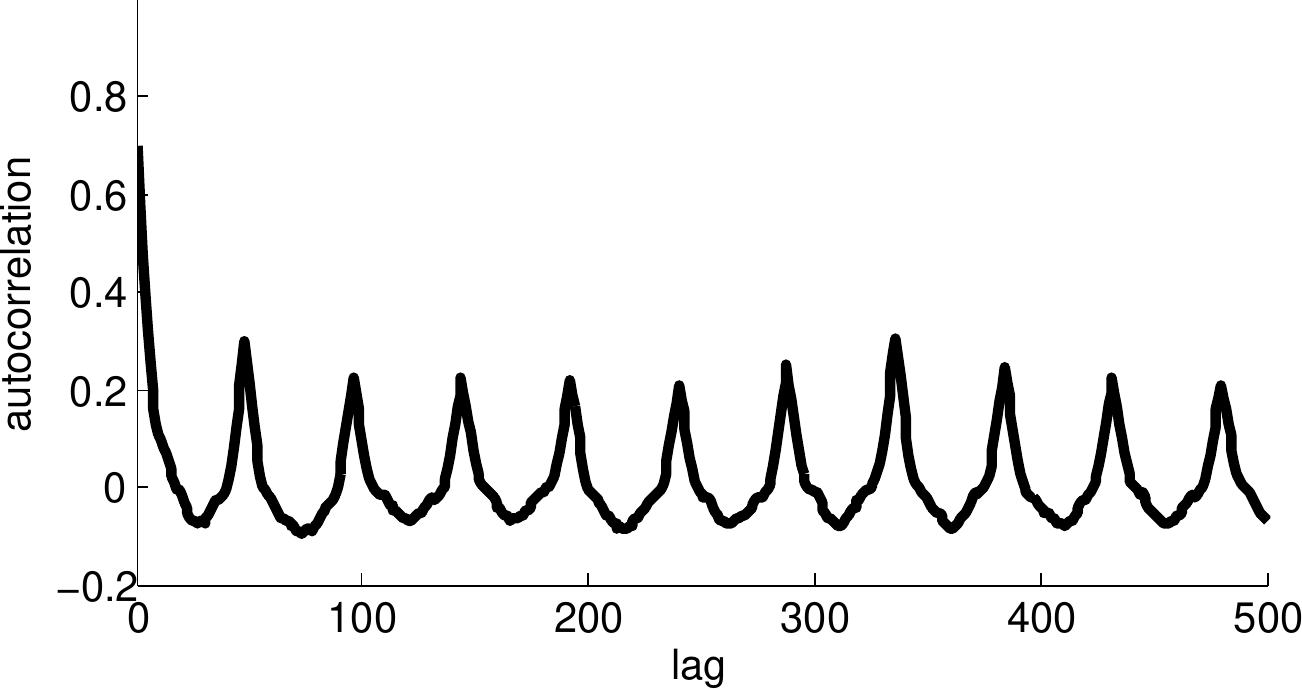}
\caption{Autocorrelation function of the Electricity labels.}
\label{fig:autocor}
\end{minipage}
\hspace{0.5cm}
\begin{minipage}[b]{0.45\linewidth}
\centering
\includegraphics[width = 0.75\textwidth]{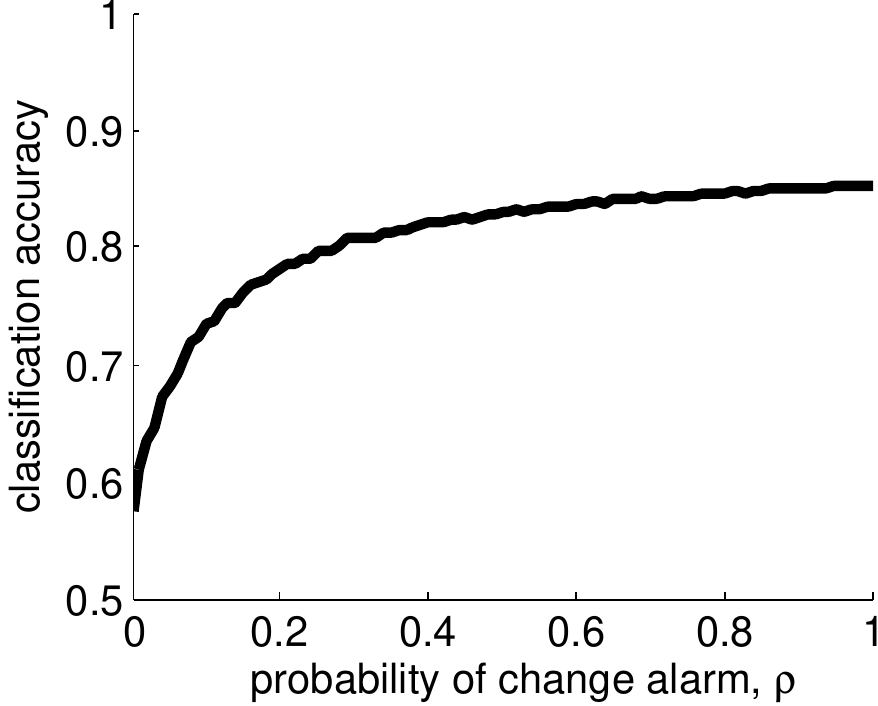}
\caption{Naive classification accuracies on the Electricity dataset.}
\label{fig:rho}
\end{minipage}
\end{figure}

This dataset has an important property not to be ignored when evaluating concept drift adaptation.
Suppose we employ a naive predictor that predicts the next label to be the same as the current label (the moving average of one).
For instance, if the price goes UP now, it predict that the next time step the price will go UP as well.
If the data was distributed independently, such a predictor would achieve $51\%$ accuracy\footnote{The probability
that two labels in a row are the same is $p(UP)^2 + p(DOWN)^2$.}.
However, if we test this naive approach on the Electricity dataset it gives much higher $85\%$ accuracy.
This happens because the labels are not independent; there are long consecutive periods of UP and long consecutive periods of DOWN.
Figure \ref{fig:autocor} plots the autocorrelation function\footnote{Autocorrelation peaks at every 48 instances (24 hours) due to the cylices of electricity consumption.} of the labels.

The problem with evaluation of adaptive classifiers on such a dataset is that we cannot be sure if a change detector (and adaptation) is working well.
Suppose we have a classifier with a worthless change detection mechanism. 
If fires a change alarm after any instance \emph{at random} with the probability $\rho$.
After firing an alarm the classifier is restarted and continues training on the most recent data.
Suppose we do not take into consideration any input data, we do not build any intelligent models just look at the labels.
If $\rho=0$, i.e. no change detection, we get the majority class (always DOWN) classifier that would achieve $58\%$ accuracy over this dataset.
If $\rho=1$, we alarm a change as often as possible, we get the moving average of one classifier.
Figure \ref{fig:rho} plots the accuracies in between.
Note that if the data was distributed independently we would get the naive accuracy $51\%$ independently of $\rho$.\\

In the appendix we report the results of testing several adaptive classifiers implemented in MOA \citep{moa} and the accuracies found in the literature on the Electricity dataset.\\
\\

In summary, the more random change alarms the classifier fires, the better the accuracy.
There change alarms are not related in detecting concept drift in any way, we are not using the input data $X$ in this experiment.
Thus, getting high accuracy on the Electricity dataset does not necessarily mean that the adaptation mechanism is good.
In such a case we recommend at least comparing the testing accuracies with the accuracy of the moving average of one.\\

This note is intended to be updated. There is a website for discussing this issue or leaving your comments \url{https://sites.google.com/site/zliobaite/about_electricity}.

\newpage
\appendix
\section*{Appendix}

Table \ref{tab:moaacc} reports classification accuracies tested with MOA\citep{moa} implementations.
We see that only LeveragingBag and AdaHoeffdingOptionTree outperform the moving average of one.
Table \ref{tab:metaacc} collects classification accuracies on the Electricity dataset as reported in published papers.

\begin{table}[h]
\caption{Accuracies of adaptive classifiers on the Electricity datasets tested with MOA.}
\begin{tabular}{lcccr}
\hline
Algorithm & Accuracy & Time & Memory & Reference\\
\hline
LeveragingBag           & 88.6 & 8.83 & 0.62 & \citep{Bifet10}\\
AdaHoeffdingOptionTree  & 86.7 & 1.61 & 0.71 & \citep{Pfahringer07}\\
{\bf moving average of one}& {\bf 85.3} & & & \\
SingleClassifierDrift EDDM&84.9& 1.00 & 0.00 &\citep{Baena06}\\
OzaBagADWIN             & 84.5 & 3.81 & 0.21 &\citep{Bifet09}\\
HoeffdingAdaptiveTree   & 83.6 & 0.97 & 0.02 &\citep{Bifet09b}\\
SingleClassifierDrift DDM &82.7& 2.23 & 0 &\citep{Gama04}\\
AccuracyUpdatedEnsemble2& 77.6 & 5.04 & 0.96 &\citep{Wang03}\\
NaiveBayes              & 74.2 & 0.30 & 0.01 &\\
AccuracyUpdatedEnsemble1& 72.8 & 7.25 & 0.74 &\citep{Wang03}\\
AccuracyWeightedEnsemble& 71.1 & 6.33 & 0.37 &\citep{Wang03}\\
AccuracyUpdatedEnsemble & 70.6 & 7.36 & 0.73 &\citep{Brzezinski11}\\
MajorityClass           & 57.5 & 0.20 & 0 \\
\hline
\end{tabular}
\label{tab:moaacc}
\end{table}

\begin{table}[h]
\caption{Accuracies of adaptive classifiers on the Electricity dataset reported in literature.}
\begin{tabular}{lcr}
\hline
Algorithm & Accuracy & Reference\\
\hline
DDM                        & 89.6* & \citep{Gama04}\\
Learn++.CDS                & 88.5 & \citep{Ditzler12}\\
KNN-SPRT                   & 88.0 & \citep{Ross12}\\
GRI                        & 88.0 & \citep{Tomczak12}\\
FISH3                      & 86.2 & \citep{Zliobaite11} \\
EDDM-IB1                   & 85.7 & \citep{Baena06}  \\
{\bf moving average of one}& {\bf 85.3} & \\
ASHT                       & 84.8 & \citep{Bifet09}\\
bagADWIN                   & 82.8 & \citep{Bifet09}\\
DWM-NB                     & 80.8 & \citep{Kolter07}\\
Local detection            & 80.4 & \citep{Gama06}  \\
Perceptron                 & 79.1 & \citep{Bifet10b} \\
ADWIN                      & 76.6 & \citep{Bifet07}\\
Prop. method               & 76.1 & \citep{Martinez11}\\
AUE                        & 74.9 & \citep{Brzezinski11}\\
Cont. $\lambda$-perc.      & 74.1 & \citep{Pavlidis11}\\
CALDS                      & 72.5 & \citep{Gomes10}\\
TA-SVM                     & 68.9 & \citep{Grinblat11} \\
\hline
\multicolumn{3}{l}{* tested on a subset} \\
\hline
\end{tabular}
\label{tab:metaacc}
\end{table}

\newpage
\bibliography{bib_electricity}

\end{document}